\documentclass[11pt,a4paper]{article}
\usepackage[dvipsnames,table]{xcolor}
\usepackage[hyperref]{eacl2021}
\usepackage{times}
\usepackage{latexsym}

\usepackage{microtype}
\aclfinalcopy 
    
\usepackage{url}
\usepackage{latexsym}

\usepackage{booktabs}
\usepackage{amsmath}
\usepackage{caption}
\usepackage{subcaption}
\usepackage{graphicx}
\usepackage{multirow}
\usepackage{multicol}
\usepackage{tablefootnote}

\graphicspath{{figures/}}

\usepackage{CJKutf8}

\usepackage[disable]{todonotes}
\presetkeys{todonotes}{color=green}{}

\newcommand{\langrank}{\textsc{LangRank}}

\newcommand{\langrankall}{\textsc{LangRank+Prag}}
\newcommand{\langvec}{\textsc{MTVec}}
\newcommand{\langvecall}{\textsc{MTVec+Prag}}
\newcommand{\markstar}{\textsuperscript{*}}

\newcommand{\Sref}[1]{\S\ref{#1}}

\newcommand{\Fref}[1]{Figure~\ref{#1}}

\newcommand{\Tref}[1]{Table~\ref{#1}}
\newcommand{\Aref}[1]{Appendix~\ref{#1}}
\newcommand\emptyfootnote[1]{%
  \begingroup
  \renewcommand\thefootnote{}\footnote{#1}%
  \addtocounter{footnote}{-1}%
  \endgroup
}

\title{Cross-Cultural Similarity Features for Cross-Lingual Transfer Learning of Pragmatically Motivated Tasks}

\author{
  Jimin Sun\textsuperscript{1*}\quad 
  Hwijeen Ahn\textsuperscript{2*}\quad  
  Chan Young Park\textsuperscript{3*}\\ 
  \textbf{Yulia Tsvetkov\textsuperscript{3}} \quad  
  \textbf{David R. Mortensen\textsuperscript{3}}\\
  \textsuperscript{1}Seoul National University, Republic of Korea\\
  \textsuperscript{2}Sogang University, Republic of Korea\\
  \textsuperscript{3}Language Technologies Institute, Carnegie Mellon University, USA\\
  {\tt jiminsun@dm.snu.ac.kr, hwijeen@sogang.ac.kr},\\
  {\tt \{chanyoun, ytsvetko, dmortens\}@cs.cmu.edu}\\
}
\date{}

\begin{document}
\maketitle
\begin{abstract}
Much work in cross-lingual transfer learning explored how to select better transfer languages for multilingual tasks, 
primarily focusing on typological and genealogical similarities between languages. 
We hypothesize that these measures of linguistic proximity 
are not enough when working with pragmatically-motivated tasks, such as sentiment analysis. 
As an alternative, we introduce three linguistic features 
that capture cross-cultural similarities that manifest in linguistic patterns and 
quantify distinct aspects of language pragmatics: 
language context-level, figurative language, and the lexification of emotion concepts. 
Our analyses show that the proposed pragmatic features 
do capture cross-cultural similarities and align well with existing work in sociolinguistics and linguistic anthropology.
We further corroborate the effectiveness of pragmatically-driven transfer in the downstream task of choosing transfer languages for cross-lingual sentiment analysis.
\end{abstract}

\section{Introduction}
\label{intro}
\emptyfootnote{*The first three authors contributed equally.}
\citet{hofstede2005cultures} defined culture as the collective mind which ``distinguishes the members of one group of people from another.''
Cultural idiosyncrasies affect and shape people's beliefs and behaviors. Linguists have particularly focused on the relationship between culture and language, revealing in qualitative case studies how cultural differences are manifested as linguistic variations \cite{doi:10.1525/aa.1977.79.2.02a00250}.

Quantifying cross-cultural similarities from linguistic patterns has largely been unexplored in NLP, with the exception of studies that focused on cross-cultural differences in word usage  \cite{garimella-etal-2016-identifying,lin-etal-2018-mining}.
In this work, we aim to quantify cross-cultural similarity, focusing on \emph{semantic} and \emph{pragmatic} differences across languages.\footnotemark{}
We devise a new distance measure between languages based on linguistic proxies of culture. We hypothesize that it can be used to select transfer languages and improve cross-lingual transfer learning, specifically in pragmatically-motivated tasks such as sentiment analysis, since expressions of subtle sentiment or emotion---such as subjective well-being \cite{smith2016does}, anger \cite{oster2019cross}, or irony \cite{karoui-etal-2017-exploring}---have been shown to vary significantly by culture.

\footnotetext{{In linguistics, \emph{pragmatics} has both a broad and a narrow sense. Narrowly, the term refers to formal pragmatics. In the broad sense, which we employ in this paper, pragmatics refers to contextual factors in language use. We are particularly concerned with cross-cultural pragmatics and finding quantifiable linguistic measures that correspond to aspects of cultural context. These measures are not the cultural characteristics that would be identified by anthropological linguists themselves but are rather intended to be measurable correlates of these characteristics.}}

We focus on three distinct aspects in the intersection of language and culture, and propose features to operationalize them. 
First, every language and culture rely on different levels of \emph{context in communication}.
Western European languages are generally considered low-context languages, whereas Korean and Japanese are considered high-context languages \cite{hall1989beyond}. 
Second, similar cultures construct and construe \emph{figurative language} similarly \cite{casas1995sociolinguistic,vulanovic2014cultural}.
Finally, \emph{emotion semantics} is similar between languages that are culturally-related \cite{Jackson1517}.
For example, in Persian, `grief' and `regret' are expressed with the same word whereas `grief' is co-lexified with `anxiety' in Dargwa.
Therefore, Persian speakers may perceive `grief' as more similar to `regret,' while Dargwa speakers may associate the concept with `anxiety.'

We validate the proposed features qualitatively, and also quantitatively by an extrinsic evaluation method.
We first analyze each linguistic feature to confirm that they capture the intended cultural patterns. We find that the results corroborate the existing work in sociolinguistics and linguistic anthropology.
Next, as a practical application of our features, we use them to rank transfer languages for cross-lingual transfer learning.
\citet{lin-etal-2019-choosing} have shown that selecting the right set of transfer languages with syntactic and semantic language-level features can significantly boost the performance of cross-lingual models. 
We incorporate our features into \citet{lin-etal-2019-choosing}'s ranking model to evaluate the new cultural features' utility in selecting better transfer languages.
Experimental results show that incorporating the features improves the performance for cross-lingual sentiment analysis, but not for dependency parsing. 
These results support our hypothesis that cultural features are more helpful when the cross-lingual task is driven by pragmatic knowledge.
\footnote{Code and data are publicly available at
\url{https://github.com/hwijeen/langrank}.} 

\section{Pragmatically-motivated Features}
\label{sec:features}
We propose three language-level features that quantify the cultural similarities across languages.

\paragraph{Language Context-level Ratio}
A language's \emph{context-level} reflects the extent to which the language leaves the identity of entities and predicates to context. For example, an English sentence \textit{Did you eat lunch?} explicitly indicates the pronoun \textit{you}, whereas the equivalent Korean sentence \text{\begin{CJK}{UTF8}{mj}점심 먹었니?\end{CJK}} (= \textit{Did eat lunch?}) omits the pronoun.
 Context-level is considered one of the distinctive attributes of a language's pragmatics in linguistics and communication studies, and if two languages have similar levels of context, their speakers are more likely to be from similar cultures \cite{korac-kakabadse-2001}.   

The language context-level ratio (\texttt{LCR}) feature approximates this linguistic quality. We compute the pronoun- and verb-token ratio, $\texttt{ptr}(l_k)$ and $\texttt{vtr}(l_k)$ for each language $l_k$, using part-of-speech tagging results. We first run language-specific POS-taggers over a large mono-lingual corpus for each language. Next, we compute \texttt{ptr} as the ratio of count of pronouns in the corpus to the count of all tokens. \texttt{vtr} is obtained likewise with verb tokens. Low $\texttt{ptr}$, $\texttt{vtr}$ values may indicate that a language leaves the identity of entities and predicates, respectively, to context. 
We then compare these values between the \textit{target language} $l_{tg}$ and \textit{transfer language} $l_{tf}$, which leads to the following definition of \texttt{LCR}: 
\begin{align*}
    \texttt{LCR-pron}(l_{tf}, l_{tg}) &= \texttt{ptr}(l_{tg})/{\texttt{ptr}(l_{tf})}\\
    \texttt{LCR-verb}(l_{tf}, l_{tg}) &= {\texttt{vtr}(l_{tg})}/{\texttt{vtr}(l_{tf})} \
\end{align*}

\paragraph{Literal Translation Quality}
Similar cultures tend to share similar figurative expressions, including idiomatic multiword expressions (MWEs) and metaphors \cite{kovecses2003, kovecses2010metaphor}. 
For example, \textit{like father like son} in English can be translated word-by-word into a similar idiom \textit{tel p\`ere tel fils} in French. 
However, in Japanese, a similar idiom \begin{CJK}{UTF8}{min}{蛙の子は蛙}\end{CJK} (\textit{Kaeru no ko wa kaeru}) ``A frog's child is a frog." cannot be literally translated.

Literal translation quality (\texttt{LTQ}) feature quantifies how well a given language pair's MWEs are preserved in literal (word-by-word) translation, using a bilingual dictionary. 
A well-curated list of MWEs is not available for the majority of languages. We thus follow an automatic extraction approach of MWEs \citep{tsvetkov-wintner-2010-extraction}. 
First, a variant of pointwise mutual information, $\text{PMI}^3$ \cite{daille1994approche} is used to extract noisy lists of top-scoring n-grams from two large monolingual corpora from different domains, and intersecting the lists filters out domain-specific n-grams and retains the language-specific top-$k$ MWEs. 
Then, a bilingual dictionary between $l_{tf}$ and $l_{tg}$ and a parallel corpus between the pair are used.
\footnote{While dictionaries and parallel corpora are not available for many languages, they are easier to obtain than the task-specific annotations of MWEs.}
For each n-gram in $l_{tg}$'s MWEs, we search for its literal translations extracted using the dictionary in parallel sentences containing the n-gram.
For any word in the n-gram, if there is a translation in the parallel sentence, we consider this as hit, otherwise as miss.
And we calculate \textit{hit ratio} as $\frac{\textit{hit}}{( \textit{hit}+ \textit{miss})}$ for each n-gram found in the parallel corpus. Finally, we average the hit ratios of all n-grams and \textit{z}-normalize over the transfer languages to obtain $\texttt{LTQ}(l_{tf},l_{tg})$.

\paragraph{Emotion Semantics Distance}
Emotion semantic distance (\texttt{ESD}) measures how similarly emotions are lexicalized across languages. This is inspired by \citet{Jackson1517} who used colexification patterns (i.e., when different concepts are expressed using the same lexical item) to capture the semantic similarity of languages. However, colexification patterns require human annotation, and existing annotations may not be comprehensive.
We extend \citet{Jackson1517}'s method by using cross-lingual word embeddings.

We define \texttt{ESD} as the average distance of emotion word vectors in transfer and target languages, after aligning word embeddings into the same space.
More specifically, we use 24 emotion concepts defined in \citet{Jackson1517} and use bilingual dictionaries to expand each concept into every other language (e.g., \textit{love} and \textit{proud} to \textit{Liebe} and \textit{stolz} in German). 
We then remove the emotion word pairs from the bilingual dictionaries, and use the remaining pairs to align word embeddings of source into the space of target languages. 
We hypothesize that if words correspond to the same emotion concept in different languages (e.g., \textit{proud} and \textit{stolz}) have similar meaning, they should be aligned to the same point despite the lack of supervision. 
However, because each language possesses different emotion semantics, emotions are scattered into different positions. We thus define \texttt{ESD} as the average cosine distance between languages:
\begin{align*}
    \texttt{ESD}(l_{tf}, l_{tg})=\sum_{e\in E}\text{cos}(\mathbf{v}_{tf,e}, \mathbf{v}_{tg,e})/{|E|}
\end{align*}
where $E$ is the set of emotion concepts and $\mathbf{v}_{tf, e}$ is the aligned emotion word vector of language $l_{tf}$.
% for emotion concept $e$.

\section{Feature Analysis}
\label{sec:feature-analysis}
In this section, we evaluate the proposed pragmatically-motivated features intrinsically. Throughout the analyses, we use 16 languages listed in \Fref{fig:lang-info} which are later used for extrinsic evaluation (\Sref{sec:exp}).

\subsection{Implementation Details}
We used multilingual word tokenizers from \href{https://www.nltk.org/api/nltk.tokenize.html}{NLTK} and \href{https://github.com/datquocnguyen/RDRPOSTagger}{RDR POS Tagger} \cite{nguyen-etal-2014-rdrpostagger} for most of the languages except for Arabic, Chinese, Japanese, and Korean, where we used \href{https://github.com/linuxscout/pyarabic}{PyArabic}, \href{https://github.com/fxsjy/jieba}{Jieba}, \href{https://github.com/neubig/kytea}{Kytea}, and \href{https://github.com/konlpy/konlpy/}{Mecab}, respectively.
For monolingual corpora, we used the news-crawl 1M corpora from \href{https://wortschatz.uni-leipzig.de/en/download}{Leipzig} \cite{goldhahn-etal-2012-building} for both \texttt{LCR} and \texttt{LTQ}. 
We used  bilingual dictionaries from \citet{choe2020word2word} and TED talks corpora \cite{qi-etal-2018-pre} for both parallel corpora and an additional monolingual corpus for \texttt{LTQ}. We focused on bigrams and trigrams and set $k$, the number of extracted MWEs, to 500. We followed \citet{lample2018word} to generate the supervised cross-lingual word embeddings for \texttt{ESD}.

\subsection{LCR and Language Context-level}

\begin{figure}[t]
    \centering
    \includegraphics[width=\linewidth]{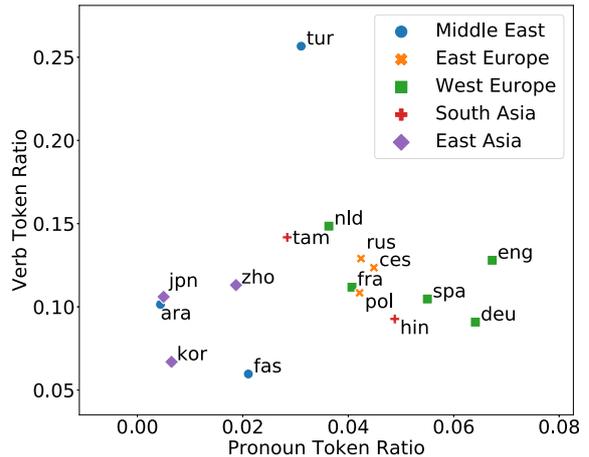}
    \caption{Plot of languages in \texttt{ptr} and \texttt{vtr} plane. Languages are color-coded according to the cultural areas defined in  \citet{doi:10.1525/aa.1977.79.2.02a00250}.
    }
    \label{fig:pos-ratio}
\end{figure}

\texttt{ptr} approximates how often discourse entities are indexed with pronouns rather than left conjecturable from context. Similarly, \texttt{vtr} estimates the rate at which predicates appear explicitly as verbs. In order to examine to which extent these features reflect context-levels, we plot languages on a two-dimensional plane where the x-axis indicates \texttt{ptr} and the y-axis indicates \texttt{vtr} in \Fref{fig:pos-ratio}.

The plot reveals a clear pattern of context-levels in different languages. Low-context languages such as German and English \cite{hall1989beyond} possess the largest values of \texttt{ptr}. On the other extreme are located Korean and Japanese with low \texttt{ptr}, which are representative of high-context languages. One thing to notice is the isolated location of Turkish with a high \texttt{vtr}. This is morphosyntactically plausible as a lot of information is expressed by the affixation to verbs in Turkish. 

\begin{figure*}
    \centering
    \begin{subfigure}[t]{.45\textwidth}
    \centering
    \includegraphics[width=\textwidth]{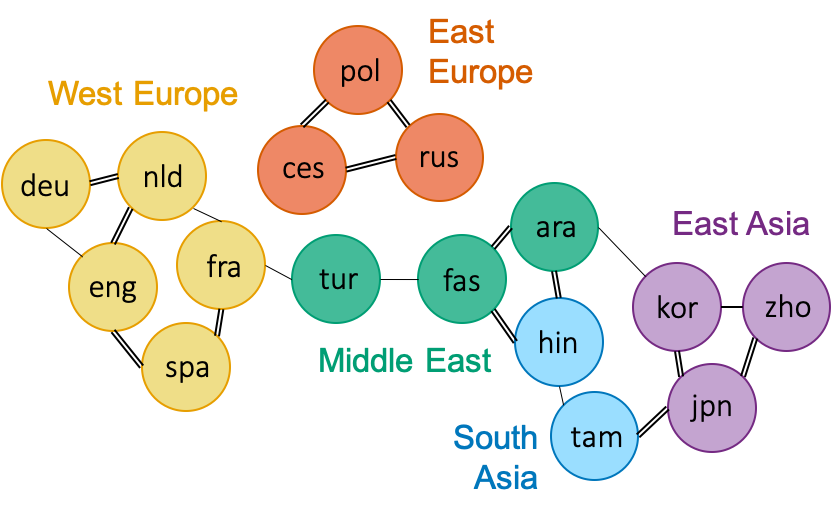}
    \caption{Network based on Emotion Semantics Distance.}
    \label{fig:esd-network}
    \end{subfigure}\quad
    \begin{subfigure}[t]{.45\textwidth}
    \centering
    \includegraphics[width=\textwidth]{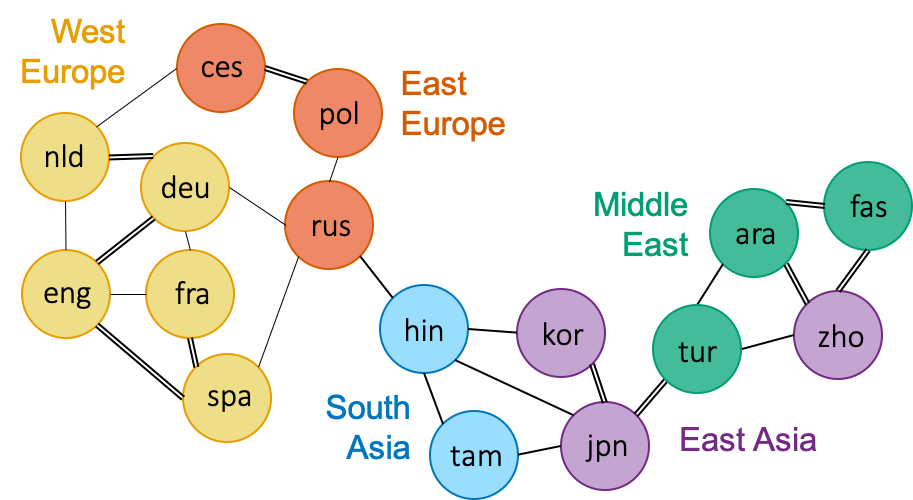}
    \caption{Network based on syntactic distance.}
    \label{fig:syn-network}
    \end{subfigure}
    \caption{
    Network of languages color-coded by their cultural areas. An edge is added between the two languages if a language is ranked in the top-2 closest languages of the other language in terms of feature value.
    }
    \label{fig:lang-network}
\end{figure*}

\subsection{LTQ and MWEs}
\texttt{LTQ} uses n-grams with high PMI scores as proxies for figurative language MWE (PMI MWEs). We evaluate the quality of selected MWEs and the resulting \texttt{LTQ} by comparing with human-curated list of figurative language MWE (gold MWEs) that are available in some languages.
We collected gold MWEs in multiple languages from Wiktionary\footnote{For example, \url{https://en.wiktionary.org/wiki/Category:English\_idioms}}.
We discarded languages with less than 2,000 phrases on the list, resulting in four languages (English, French, German, Spanish) for analysis.

First, we check how many PMI MWEs are actually in the gold MWEs. Out of the top-500 PMI bigrams and trigrams, 19.0\% of bigrams and 3.8\% of trigrams are included in the gold MWE list (averaged over four languages).  
For example, the trigrams in the PMI MWEs, \textit{keep an eye} and \textit{take into account}, are considered to be in the gold MWEs as \textit{keep an eye peeled} and \textit{take into account} are in the list.
The seemingly low percentages are reasonable, regarding that the PMI scores are designed to extract collocations patterns rather than figurative languages themselves.

Secondly, to validate using PMI MWEs as proxies, we compare the \texttt{LTQ} of PMI MWEs with the \texttt{LTQ} using gold MWEs.
Specifically, we obtained the \texttt{LTQ} scores of each language pair with target languages limited to the four European languages mentioned above. 
Then for each target language, we measured Pearson correlation coefficient between the two \texttt{LTQ} scores based on the two MWE lists.
The average coefficient was 0.92, which indicates a strong correlation between the two resulting \texttt{LTQ} scores, and thus justifies using PMI MWEs for all other languages.

\subsection{ESD and Cultural Grouping}
We investigate what is carried by \texttt{ESD} by visualizing and looking at the nearest neighbors of emotion vectors.\footnote{A visualization demo of emotion vectors can be found at
\url{https://bit.ly/emotion\_vecs}.}
\citet{Jackson1517} used word colexification patterns to reveal that the same emotion concepts cluster with different emotions according to the language family they belong to. 
For instance, in Tai-Kadai languages, \textit{hope} appears in the same cluster as \textit{want} and \textit{pity}, while \textit{hope} associates with \textit{good} and \textit{love} in the Nakh-Daghestanian language family. 
Our results derived from \texttt{ESD} do not rely on colexification patterns, but also support this finding.
The nearest neighbors of the Chinese word for \textit{hope} was \textit{want} and \textit{pity}, while they were found as \textit{love} and \textit{joy} for \textit{hope} in Arabic.

In \Fref{fig:lang-network}, we compare \texttt{ESD} to the syntactic distance between languages by constructing two networks of languages based on each feature. \Fref{fig:esd-network} uses \texttt{ESD} as reference while \Fref{fig:syn-network} uses the syntactic distance from the URIEL database \cite{littell-etal-2017-uriel}. 
Each node represents a language, color-coded by its cultural area.
For each language, we sort the other languages according to the distance value. When a language is in the list of top-$k$ closest languages, we draw an edge between the two. We set $k=2$.

We see that languages in the same cultural areas tend to form more cohesive clusters in \Fref{fig:esd-network} compared to \Fref{fig:syn-network}. The portion of edges \emph{within} the cultural areas is $76\%$ for \texttt{ESD} while it is $59\%$ for syntactic distance. These results indicate that \texttt{ESD} effectively extracts linguistic information that aligns well with the commonly shared perception of cultural areas.

\subsection{Correlation with Geographical Distance}
\label{ssec:geo}
Regarding the language clusters in \Fref{fig:esd-network}, some may suspect that geographic distance can substitute the pragmatically-inspired features. For Chinese, Korean and Japanese are the closest languages by \texttt{ESD}, which can also be explained by their geographical proximity. Do our features add additional pragmatic information, or can they simply be replaced by geographical distance?

To verify this speculation, we evaluate Pearson's correlation coefficient of each pragmatic feature value with geographical distance from URIEL. The feature with the strongest correlation was \texttt{ESD} $(r{=}0.4)$. The least correlated was \texttt{LCR-verb} $(r{=}0.03)$, followed by \texttt{LCR-pron} $(r{=}0.17)$ and \texttt{LTQ} $(r{=}{-}0.31)$\footnotemark. The results suggest that the pragmatic features contain extra information that cannot be subsumed by geographic distance.
\footnotetext{When two languages are more similar, LTQ is higher whereas geographic distance is smaller.}

\section{Extrinsic Evaluation: Ranking Transfer Languages}
\label{sec:extrinsic_eval}

To demonstrate the utility of our features, we apply them to a \emph{transfer language ranking} task for cross-lingual transfer learning.
We first present the overall task setting, including the datasets and models used for the two cross-lingual tasks.
Next, we describe the transfer language ranking model and its evaluation metrics.

\subsection{Task Setting}
\label{sec:problem}

\begin{figure}[t]
    \centering
    \includegraphics[width=0.9\linewidth]{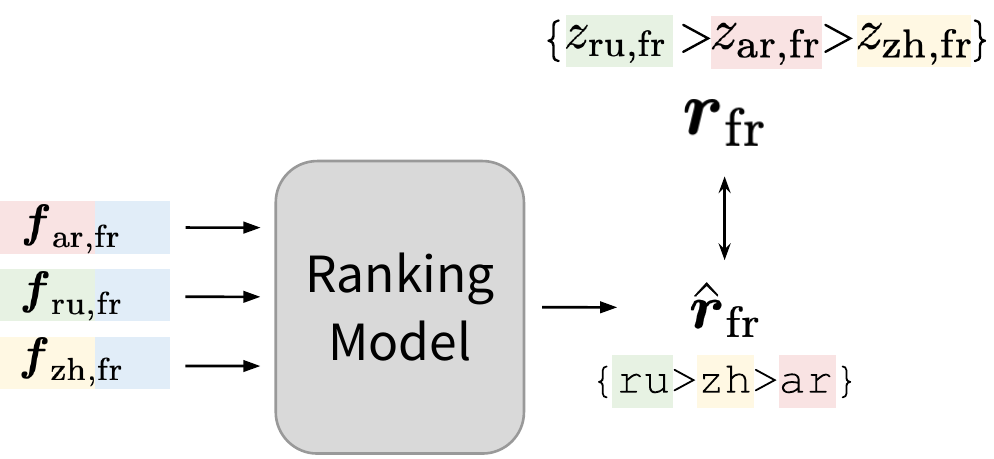}
    \caption{
    Illustration of transfer language ranking problem when the target language is French (fr) and there are three available transfer languages: Arabic (ar), Russian (ru), and Chinese (zh).
    The output ranking $\hat{\boldsymbol{r}}_{\text{fr}}$ is compared to the ground truth ranking $\boldsymbol{r}_{\text{fr}}$ which is determined by the zero-shot performance $\boldsymbol{z}$ of cross-lingual models.
    }
    \label{fig:eval}
\end{figure}

We define our task as the \emph{language ranking} problem:
given the target language $l_{tg}$, we want to rank a set of $n$ candidate transfer languages $\mathcal{L}_{tf}{=}\{l_{tf}^{(1)}, \ldots , l_{tf}^{(n)}\}$ by their usefulness when transferred to $l_{tg}$, which we refer to as \emph{transferability} (illustrated in \Fref{fig:eval}).
The effectiveness of cross-lingual transfer is often measured by evaluating the joint training or zero-shot transfer performance \cite{wu-dredze-2019-beto,schuster-etal-2019-cross}. 
In this work, we quantify the effectiveness as the zero-shot transfer performance, following \citet{lin-etal-2019-choosing}. 
Our goal is to train a model that ranks available transfer languages in $\mathcal{L}_{tf}$ by their transferability for a target language $l_{tg}$.

To train the ranking model, we first need to find the ground-truth transferability rankings, which operate as the model's training data.
We evaluate the zero-shot performance $z_{tf,tg}$ by training a task-specific cross-lingual model solely with transfer language $l_{tf}$ and testing on $l_{tg}$. 
After evaluating $z_{tf,tg}$ for each candidate transfer language in $\mathcal{L}_{tf}$, we obtain the optimal ranking of languages $\boldsymbol{r}_{tg}$ by sorting languages according to the measured $z_{tf,tg}$. 
Note that $\boldsymbol{r}_{tg}$ also depends on downstream task.

Next, we train the language ranking model. 
The ranking model predicts the transfer ranking of candidate languages. Each source, target pair $(l_{tf},l_{tg})$ is represented as a vector of language features $\boldsymbol{f}_{tf,tg}$, which may include phonological similarity, typological similarity, word-overlap to name a few. The ranking model takes $\boldsymbol{f}_{tf,tg}$ of every $l_{tf}$ as input, and predicts the transferability ranking $\boldsymbol{\widehat{r}_{tg}}$.
Using $\boldsymbol{r}_{tg}$ from the previous step as training data, the model learns to find optimal transfer languages based on $\boldsymbol{f}_{tf,tg}$.
The trained model can either be used to select the optimal set of transfer languages, or to decide which language to additionally annotate during the data creation process.

\subsection{Task \& Dataset}
\label{sec:data}

\begin{figure}[t]
    \centering
    \includegraphics[width=0.7715\linewidth]{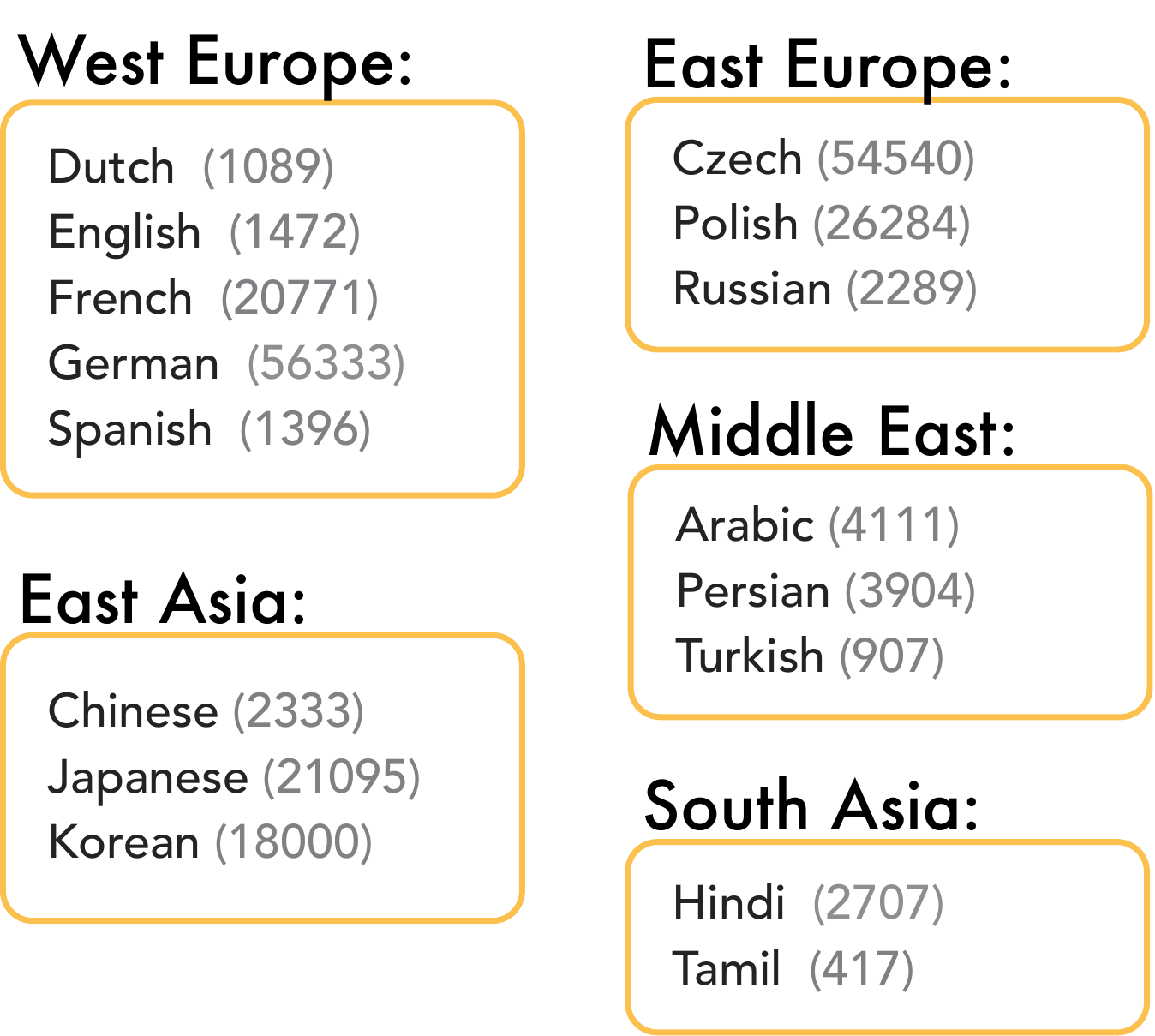}
    \caption{Languages used throughout the experiments are grouped by their cultural areas \cite{doi:10.1525/aa.1977.79.2.02a00250}. The numbers indicate the size of each dataset.}
    \label{fig:lang-info}
\end{figure}

We apply the proposed features to train a ranking model for two distinctive tasks: multilingual sentiment analysis (SA) and multilingual dependency parsing (DEP). 
The tasks are chosen based on our hypothesis that high-order information such as pragmatics would assist sentiment analysis while it may be less significant for dependency parsing, where lower-order information such as syntax is relatively stressed.

\paragraph{SA}
As there is no single sentiment analysis dataset covering a wide variety of languages, we collected various review datasets from different sources.\footnote{Details are provided in \Aref{appendix:dataset}. Note that the difference in domain and label distribution of data can also affect the transferability, and a related discussion is in \Sref{ssec:controlled_experiment}}
All samples are labeled as either positive or negative. In case of datasets rated with a five-point Likert scale, we mapped 1--2 to negative and 4--5 to positive. We settled on a dataset consist of 16 languages categorized into five distinct cultural groups: \text{West Europe}, \text{East Europe}, \text{East Asia}, \text{South Asia}, and \text{Middle East} (\Fref{fig:lang-info}).

\paragraph{DEP}
To compare the effectiveness of the proposed features on syntax-focused tasks, we chose datasets of the same set of 16 languages from Universal Dependencies v2.2 \cite{11234/1-2837}.

\subsection{Task-Specific Cross-Lingual Models}
\label{sec:eval}

\paragraph{SA}
Multilingual BERT (mBERT) \cite{devlin-etal-2019-bert}, a multilingual extension of BERT pretrained with 104 different languages, has shown strong results in various text classification tasks in cross-lingual settings \cite{10.1007/978-3-030-32381-3_16,xu-etal-2019-bert, li-etal-2019-exploiting}. 
We use mBERT to conduct zero-shot cross-lingual transfer and to extract optimal transfer language rankings: fine-tune mBERT on transfer language data and test it on target language data. The performance is measured by the macro F1 score on the test set.

\paragraph{DEP}
We adopt the setting from \citet{DBLP:journals/corr/abs-1811-00570} to perform cross-lingual zero-shot transfer. 
We train deep biaffine attentional graph-based models \cite{DBLP:journals/corr/DozatM16} which achieved state-of-the-art performance in dependency parsing for many languages.
The performance is evaluated using labeled attachment scores (LAS).

\subsection{Ranking Model \& Evaluation}
\paragraph{Ranking Model}
For the language ranking model, we employ gradient boosted decision trees, LightGBM \cite{NIPS2017_6907}, which is one of the state-of-the-art models for ranking tasks.\footnote{More details on the cross-lingual models, ranking model, and their training can be found in \Aref{appendix:task_models}.}

\paragraph{Ranking Evaluation Metric}
We evaluate the ranking models' performance with two standard metrics for ranking tasks: Mean Average Precision (MAP) and Normalized Discounted Cumulative Gain at position $p$ (NDCG@$p$) \cite{jarvelin2002cumulated}.
While MAP assumes a binary concept of relevance, NDCG is a more fine-grained measure that reflects the ranking positions. The \emph{relevant} languages for computing MAP are defined as the top-$k$ languages in terms of zero-shot performance in the downstream task. In our experiments, we set $k$ to 3 for MAP. Similarly, we use NDCG@3.

We train and evaluate the model using leave-one-out cross-validation: where one language is set aside as the test language while other languages are used to train the ranking model. Among the training languages, each language is posited in turn as the \emph{target} language while others are the \emph{transfer} languages.

\section{Experiments}
\label{sec:exp}

\subsection{Baselines}
\paragraph{\langrank} 
\langrank{} \cite{lin-etal-2019-choosing} uses 13 features to train the ranking model:
The dataset size in transfer language (\texttt{tf\_size}), target language (\texttt{tg\_size}), and the ratio between the two (\texttt{ratio\_size}); Type-token-ratio (\texttt{ttr}) which measures lexical diversity and \texttt{word\_overlap} for lexical similarity between a pair of languages; various distances between a language pair from the URIEL database (geographic \texttt{geo}, genetic \texttt{gen},  inventory \texttt{inv}, syntactic \texttt{syn},  phonological \texttt{phon} and featural \texttt{feat}).

\paragraph{\langvec}
\citet{malaviya-etal-2017-learning} proposed to learn a language representation while training a neural machine translation (NMT) system in a simliar fashion to \citet{johnson-etal-2017-googles}. During training, a language token is prepended to the source sentence and the learned token's embedding becomes the language vector.
\citet{bjerva2019language} has shown that such language representations contain various types of linguistic information ranging from word order to typological information.
We used the one released by \citet{malaviya-etal-2017-learning} which has the dimension of 512.

\subsection{Individual Feature Contribution}

\begin{table}[t]
\centering
\resizebox{\columnwidth}{!}{
\begin{tabular}{@{}l|ll|ll@{}}
\toprule
             & \multicolumn{2}{c}{SA}                               & \multicolumn{2}{c}{DEP}                              \\
             & \multicolumn{1}{c}{MAP} & \multicolumn{1}{c}{NDCG} & \multicolumn{1}{c}{MAP} & \multicolumn{1}{c}{NDCG} \\ \midrule
\langrank     &   71.3  &        86.5     &     \textbf{63.0}     &              \textbf{82.2}\\
\langrankall          &        \textbf{76.0}                 &  
\textbf{90.9}  &     61.7     &        80.5             \\
\rowcolor{lightgray!20}
\hspace{0.3cm}- \texttt{LCR}     &  75.0 &        88.3     & 60.3    &    79.6\\
\rowcolor{lightgray!20}
\hspace{0.3cm}- \texttt{LTQ}     &  72.4 &        89.3     & 63.1\markstar&    81.3\markstar \\
\rowcolor{lightgray!20}
\hspace{0.3cm}- \texttt{ESD}     &  77.7\markstar &        92.1\markstar & 58.2    &    78.5\\\midrule
\langvec    &  71.1  &        89.5     &     43.0     &              69.7\\
\langvecall     &  \textbf{74.3} &        \textbf{90.8}     & \textbf{49.7}    &    \textbf{74.8} \\ 
\rowcolor{lightgray!20}
\hspace{0.3cm}- \texttt{LCR}     &  72.9 &        90.1     & 54.1\markstar    &    76.3\markstar\\
\rowcolor{lightgray!20}
\hspace{0.3cm}- \texttt{LTQ}     &  71.2 &        89.0     & 53.0\markstar &    78.6\markstar \\
\rowcolor{lightgray!20}
\hspace{0.3cm}- \texttt{ESD}     &  73.1 &        90.7     & 45.3    &    73.9\\
\bottomrule
\end{tabular}
}
\caption{Evaluation results of our features (\textsc{Prag}) added to each baseline. The higher scores are \textbf{boldfaced}. Rows in \colorbox{lightgray!20}{gray} indicate ablation studies. \newline * is marked when improvements are made compared to \langrankall{}, \langvecall{} respectively.}
\label{tab:res_individual}
\end{table}

We first look into whether the proposed features are helpful in ranking transfer languages for sentiment analysis and dependency parsing (\Tref{tab:res_individual}).
We add all three features (\textsc{Prag}) to the two baseline features (\langrank, \langvec) and compare the performance in the two tasks.
Results show that our features improve both baselines in SA, implying that the pragmatic information captured by our features is helpful for discerning the subtle differences in sentiment among languages.

In the case of DEP, including our features brings inconsistent results to performance. The features help the performance of \langvec{} while they deteriorate the performance of \langrank{}. Although some performance increase was observed when applied to \langvec{}, the performance of \langvec{} in DEP remains extremely poor. These conflicting trends suggest that pragmatic information is not crucial to less pragmatically-driven tasks, represented as dependency parsing in our case.

The low performance of \langvec{} in DEP is noticeable as \langvec{} is generally believed to contain a significant amount of syntactic information, with much higher dimensionality than \langrank{}. It also suggests the limitation of using distributional representations as language features; their lack of interpretability makes it difficult to control the kinds of information used in a model.

We additionally conduct ablation studies by removing each feature from the \textsc{+Prag} models to examine each feature's contribution. 
The SA results show that \texttt{LCR} and \texttt{LTQ} significantly contribute to overall improvements achieved by adding our features, while \texttt{ESD} turns out to be less helpful. Sometimes, removing \texttt{ESD} resulted in a better performance.
In contrast, the results of DEP show that \texttt{ESD} consistently made a significant contribution, and \texttt{LCR} and \texttt{LTQ} were not useful.
The results imply that the emotion semantics information of languages is surprisingly not useful in sentiment analysis, but more so in dependency parsing.

\subsection{Group-wise Contribution} 

The previous experiment suggests that the same pragmatic information can be helpful to different extents depending on the downstream task. 
We further investigate to what extent each kind of information is useful to each task by conducting group-wise comparisons. 
To this end, we group the features into five categories: {Pretrain-specific}, {Data-specific}, {Typology}, {Geography}, {Orthography}, and {Pragmatic}.
Pretrain-specific features cover factors that may be related to the performance of pretrained language models used in our task-specific cross-lingual models. Specifically, we used the size of the Wikipedia training corpus of each language used in training mBERT.\footnote{\url{https://meta.wikimedia.org/wiki/List\_of\_Wikipedias}}
Note that we do not measure this feature group's performance on DEP as no pretrained language model was used in DEP. 
Data-specific features include \texttt{tf\_size}, \texttt{tg\_size}, and \texttt{ratio\_size}.
Typological features include \texttt{geo}, \texttt{syn}, \texttt{feat}, \texttt{phon}, and \texttt{inv} distances. Geography includes \texttt{geo} distance in isolation.
Orthographic feature is the \texttt{word\_overlap} between languages.
Finally, the Pragmatic group consists of \texttt{ttr} and the three proposed features, \texttt{LCR}, \texttt{LTQ}, and \texttt{ESD}. 
\texttt{ttr} is included in Pragmatic as \citet{richards1987type} have suggested that it encodes a significant amount of cultural information.

\Tref{tab:res_group} reports the performance of ranking models trained with the respective feature category. 
Interestingly, the two tasks showed significantly different results; the Pragmatic group showed the best performance in SA while the Typology group outperformed all other groups in DEP. 
This again confirms that the features indicating cross-lingual transferability differ depending on the target task.
Although the Pretrain-specific features were more predictive than the Geography and Orthography features it was not as helpful as the Pragmatic features.

\begin{table}[t]
\centering
\resizebox{0.95\columnwidth}{!}{
\begin{tabular}{@{}l|cc|cc@{}}
\toprule 
              & \multicolumn{2}{c}{SA}                               & \multicolumn{2}{c}{DEP}                              \\
              & \multicolumn{1}{c}{MAP} & \multicolumn{1}{c}{NDCG} & \multicolumn{1}{c}{MAP} & \multicolumn{1}{c}{NDCG} \\ \midrule
Pretrain-specific & 39.0 & 55.5 & - & - \\
Data-specific &  68.0        &      85.4    & 37.2                           &          55.0           \\
Typology      &  44.9        &      60.7    & \textbf{58.0}                 &          \textbf{79.8}   \\
Geography     &  24.9       &      55.0    & 32.3                           &          65.1           \\
Orthography   &  34.2        &      56.6    & 35.5                           &          60.5           \\
Pragmatic     & \textbf{73.0}& \textbf{88.0}& 46.5                           &          71.8          \\
\bottomrule
\end{tabular}
}
\caption{Ranking performance using each feature group as input to the ranking model.}
\label{tab:res_group}
\end{table}

\subsection{Controlling for Dataset Size}
\label{ssec:controlled_experiment}
The performance of cross-lingual transfer depends not only on the cultural similarity between transfer and target languages but also on other factors, including dataset size and label distributions. 
Although our model already accounts for the dataset size to some extent by including \texttt{tf\_size} as input, we conduct a more rigorous experiment to better understand the importance of cultural similarity in language selection.
Specifically, we control the data size by down-sampling all SA data to match both the size and label distribution of the second smallest Turkish dataset.\footnote{The size of the smallest language (Tamil; 417 samples) was too small to train an effective model.} 
We then trained two ranking models equipped with different sets of features: \langrank{} and \langrankall{}.

In terms of languages, we focus on a setting where Turkish is the target and Arabic, Japanese and Korean are the transfer languages. This is a particularly interesting set of languages because the source languages are similar/dissimilar to Turkish in different aspects; Korean and Japanese are typologically similar to Turkish, yet in cultural terms, Arabic is more similar to Turkish.

In this controlled setting, the ground-truth ranking reveals that the optimal transfer language among the three is Arabic, followed by Korean and Japanese. It indicates the important role of cultural resemblance in sentiment analysis which encapsulates the rich historical relationship shared between Arabic- and Turkish-speaking communities. \langrankall{} chose Arabic as the best transfer language, suggesting that the imposed cultural similarity information from the features helped the ranking model learn the cultural tie between the two languages. 
On the other hand, \langrank{} ranked Japanese the highest over Arabic, possibly because the provided features mainly focus on typological similarity over cultural similarity.

\section{Related Work}
\label{sec:rw}

\paragraph{Quantifying Cross-cultural Similarity}
A few recent work in psycholinguistics and NLP have aimed to measure cultural differences, mainly from word-level semantics. 
\citet{lin-etal-2018-mining} suggested a cross-lingual word alignment method that preserves the cultural, social context of words. They derive cross-cultural similarity from the embeddings of a bilingual lexicon in the shared representation space.
\citet{Thompson_Roberts_Lupyan_2018} computed similarity by comparing the nearest neighborhood of words in different languages, showing that words in some domains (e.g., time, quantity) exhibit higher cross-lingual alignment than other domains (e.g., politics, food, emotions). \citet{Jackson1517} represented each language as a network of emotion concepts derived from their colexification patterns and measured the similarity between networks.

\paragraph{Auxiliary Language Selection in Cross-lingual tasks}
There has been active work on leveraging multiple languages to improve cross-lingual systems \cite{neubig-hu-2018-rapid,ammar-etal-2016-many}. Adapting auxiliary language datasets to the target language task can be practiced through either language-selection or data-selection. Previous work on language-selection mostly relied on leveraging syntactic or semantic resemblance between languages (e.g. ngram overlap) to choose the best transfer languages \cite{zoph-etal-2016-transfer,wang-neubig-2019-target}. 
Our approach extends this line of work by leveraging cross-cultural pragmatics, an aspect that has been unexplored by prior work. 

\section{Future Directions}
\paragraph{Typology of Cross-cultural Pragmatics}
The features proposed here provide three dimensions in a provisional quantitative cross-linguistic typology of pragmatics in language. Having been validated, both intrinsically and extrinsically, they can be used in studies as a stand-in for cross-cultural similarity. They also open a new avenue of research, raising questions about what other quantitative features of language are correlates of cultural and pragmatic difference.

\paragraph{Model Probing} 
Fine-tuning pretrained models to downstream tasks has become the de facto standard in various NLP tasks, and their success has promoted the development of their multilingual extensions \cite{devlin-etal-2019-bert,lample2019cross}.
While the performance gains from these models are undeniable, their learning dynamics remain obscure. This issue has prompted various probing methods designed to test what kind of linguistic information the models retain, including syntactic and semantic knowledge \cite{conneau-etal-2018-cram,liu-etal-2019-linguistic,ravishankar-etal-2019-multilingual,DBLP:conf/iclr/TenneyXCWPMKDBD19}. Similarly, our features can be employed as a touchstone to evaluate a model's knowledge in cross-cultural pragmatics. Investigating how different pretraining tasks affect the learning of pragmatic knowledge will also be an interesting direction of research.

\section{Conclusion}
In this work, we propose three pragmatically-inspired features that capture cross-cultural similarities that arise as linguistic patterns: language context-level ratio, literal translation quality, and emotion semantic distance.
Through feature analyses, we examine whether our features can operate as valid proxies of cross-cultural similarity.
From a practical standpoint, the experimental results show that our features can help select the best transfer language for cross-lingual transfer in pragmatically-driven tasks, such as sentiment analysis.

\section*{Acknowledgements}
The authors are grateful to the anonymous reviewers for their invaluable feedback. 
This material is based upon work supported by the National Science Foundation under Grant No.~IIS2007960. 
We would also like to thank Amazon for providing AWS credits.

\bibliographystyle{acl_natbib}
\bibliography{eacl2021}

\clearpage
\begin{appendix}

\section{Dataset for Sentiment Analysis}
\label{appendix:dataset}

\begin{table}[h]
    \centering
    \begin{tabular}{@{}p{6cm} cccc@{}}
    \toprule
    \multicolumn{1}{p{6cm}}{Dataset} &
    \multicolumn{1}{l}{Languages} & \multicolumn{1}{l}{Domain} & \multicolumn{1}{l}{Size} & \multicolumn{1}{l}{POS/NEG} \\ \midrule
    \multirow{7}{*}{\href{http://alt.qcri.org/semeval2016/task5/}{\parbox{8cm}{SemEval-2016 Aspect Based \\ Sentiment Analysis}}} & Chinese & electronics & 2333 & 1.53\\
    &Arabic& hotel & 4111 & 1.54\\
    &English & restaurant & 1472 & 2.14\\
    &Dutch & restaurant & 1089 & 1.43\\
    &Spanish & restaurant & 1396 & 2.82\\
    &Russian & restaurant & 2289 & 3.81\\
    &Turkish& restaurant & 907 & 1.32\\ \midrule
    \href{https://arxiv.org/ftp/arxiv/papers/1801/1801.07737.pdf}{SentiPers}&Persian& product & 3904 & 1.8\\
    \midrule
    \multirow{3}{*}{\parbox{6cm}{Amazon Customer Reviews}}&French & product & 20771 & 8.0\\
    & German & product & 56333 & 6.56\\
    &Japanese& product & 21095 & 8.05\\\midrule
    \href{http://nlp.kiv.zcu.cz/research/sentiment}{CSFD CZ}&Czech& movie & 54540 & 1.04\\\midrule
    \parbox{6cm}{\href{https://github.com/e9t/nsmc}{Naver Sentiment Movie Corpus}}&Korean& movie & 18000 & 1.0\\\midrule
    \href{https://www.kaggle.com/sudalairajkumar/tamil-nlp}{Tamil Movie Review Dataset}&Tamil& movie & 417 & 0.48\\\midrule
    \href{http://clip.ipipan.waw.pl/PolEval?action=AttachFile&do=view&target=poleval-2017-task-1ab-gold-2.0-tei.tar.gz}{PolEval 2017}&Polish& product & 26284 & 1.38\\\midrule
    \href{http://www.lrec-conf.org/proceedings/lrec2016/pdf/698\_Paper.pdf}{Aspect based Sentiment Analysis}&Hindi& product & 2707 & 3.22\\
    \bottomrule
    \end{tabular}
    \caption{Datasets for sentiment analysis.}
    \label{tab:datasets}
\end{table}

\section{Task-Specific Models Details}
\label{appendix:task_models}

\paragraph{SA Cross-lingual Model}
We performed supervised fine-tuning of multilingual BERT (mBERT) \cite{devlin-etal-2019-bert} for the sentiment analysis task, as the model showed strong results in various text classification tasks in cross-lingual settings \cite{10.1007/978-3-030-32381-3_16,xu-etal-2019-bert, li-etal-2019-exploiting}. 
mBERT is pretrained with 104 different languages, including the 16 languages we used throughout our experiment. 
We used a concatenation of mean and max pooled representations from mBERT's penultimate layer, as it outperformed the standard practice of using the last layer's \texttt{[CLS]} token. The representation was passed to a fully connected layer for prediction. To extract optimal transfer rankings, we conducted zero-shot transfer with mBERT: fine-tuned mBERT on transfer language data and tested it on target language data. 

\paragraph{Ranking Model}
We used LightGBM \cite{NIPS2017_6907} with LambdaRank \cite{NIPS2006_2971} algorithm. The model consists of 100 decision trees with 16 leaves each, and it was trained with the learning rate of 0.1. We optimized NDCG to train the model \cite{jarvelin2002cumulated}.

\end{appendix}

\end{document}